\documentclass[letterpaper]{article} 
\usepackage{aaai2026}  
\usepackage{times}  
\usepackage{helvet}  
\usepackage{courier}  
\usepackage[hyphens]{url}  
\usepackage{graphicx} 
\usepackage{natbib}  
\usepackage{caption} 
\usepackage{amsmath}
\usepackage{algorithm}
\usepackage{algorithmic}
\usepackage{graphicx}
\usepackage{inconsolata}
\usepackage{multirow}
\usepackage{subcaption}
\usepackage{booktabs}
\usepackage{colortbl}
\usepackage{color, xcolor} 
\usepackage{amssymb}
\usepackage{enumitem}
\usepackage{newfloat}
\usepackage{listings}
\DeclareCaptionStyle{ruled}{labelfont=normalfont,labelsep=colon,strut=off} 
\floatstyle{ruled}
\newfloat{listing}{tb}{lst}{}
\floatname{listing}{Listing}
\title{Uncovering and Mitigating Transient Blindness in Multimodal Model Editing} 

\author{
    Xiaoqi Han\textsuperscript{\rm 1}, 
    Ru Li\textsuperscript{\rm 1$^{*}$}, 
    Ran Yi\textsuperscript{\rm 2}, 
    Hongye Tan\textsuperscript{\rm 1},  \\
    Zhuomin Liang\textsuperscript{\rm 1}, 
    V\'ictor Guti\'errez-Basulto\textsuperscript{\rm 3}, 
    Jeff Z. Pan\textsuperscript{\rm 4\thanks{Ru Li and Jeff Z. Pan are the corresponding authors.} }
}
\affiliations{
    \textsuperscript{\rm 1} School of Computer and Information Technology, Shanxi University, China \\
    \textsuperscript{\rm 2} Department of Computer Science and Engineering, Shanghai Jiao Tong University, China \\
    \textsuperscript{\rm 3} School of Computer Science and Informatics, Cardiff University, UK \\
    \textsuperscript{\rm 4}  ILCC, School of Informatics, University of Edinburgh, UK
}

\usepackage{bibentry}

\begin{document}

\maketitle

\begin{abstract}
Multimodal Model Editing (MMED) aims to 
correct erroneous knowledge in multimodal models. 
Existing evaluation methods, adapted from textual model editing, overstate success by relying on     
low-similarity or random inputs, 
obscure overfitting.  
We propose a comprehensive locality evaluation framework, covering 
three key dimensions: \emph{random-image locality, no-image locality,} and \emph{consistent-image locality}, 
operationalized through seven distinct data types, enabling a detailed and structured analysis of multimodal edits. 
We introduce De-VQA, a \emph{dynamic evaluation for visual question answering}, 
uncovering a phenomenon we term \emph{transient blindness},  overfitting to edit-similar text while ignoring visuals. 
Token analysis shows edits disproportionately affect textual tokens. 
We propose \emph{locality-aware adversarial losses} 
to balance 
cross-modal  representations. Empirical results demonstrate that our approach consistently outperforms existing baselines, reducing transient blindness and improving locality   by   17\% on average. 
\end{abstract}

\begin{links}
    \link{Code and Appedix with Additional Results}{https://github.com/sev777/DE-VQA}
\end{links}

\section{Introduction}
{The rapid advancement of large language models (LLMs), such as ChatGPT and Deepseek \citep{Transformer,openai2024gpt4}, has driven their widespread adoption as sources of factual knowledge~\citep{HZLV+2025,YWCL+2025,ZLP2025} for downstream tasks~\citep{PRKS+2023},  and 
multimodal LLMs have further extended these capabilities to vision-language  tasks,  
with strong performance in cross-modal understanding \citep{openai2024gpt4,blip2,minigpt4}.
However, a critical challenge for 
multimodal LLMs is knowledge obsolescence, 
making regular updates essential to maintain 
accuracy. To address this, \emph{model editing} \citep{enn,ke,mend,RASE} has emerged as an efficient solution to 
correct factual inaccuracies without 
costly retraining.
}

\begin{figure}[t]
\centering
  \includegraphics[scale=0.73]{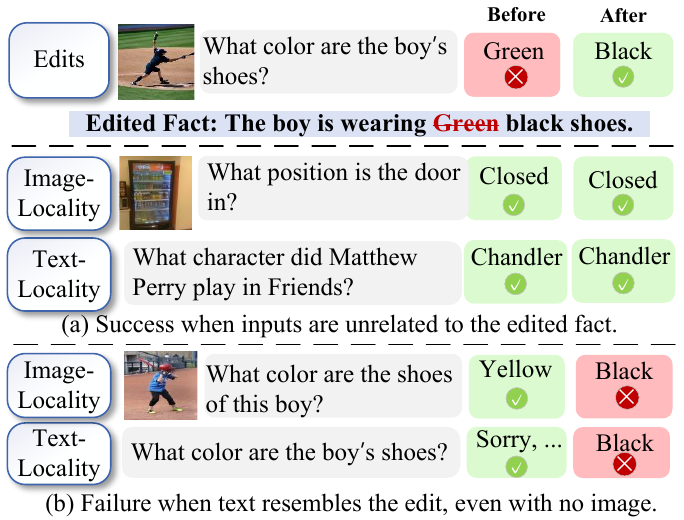} 
  \caption{{Current locality evaluation focuses only on low-similarity data (a), while the edited model   fail on high-similarity cases 
  (b).}}
  \label{example}
\end{figure}

Multimodal model editing (MMED)  specifically targets 
updating a multimodal model's predictions for specific image-text pairs while preserving   performance on unrelated inputs.
Recent works by \citet{MMED} have introduced dedicated datasets and adapted evaluation metrics for this task.
These metrics address two core aspects:
(1) Factual updating, assessed through dimensions like \textit{reliability}, \textit{generalization} to comprehensively measure an edit's effectiveness;
(2) Side-effect evaluation, referred to as \textit{locality}, where current methods simply sample random text or image-text questions  (see Figure~\ref{example} (a)).
Building on this foundation, recent methods~\cite{huang2024vlkeb,ma2025comprehendedit,dummke} have proposed more challenging factual evaluation protocols to better assess an editor's editing capacity. 
However, they largely overlook the locality aspect, an essential factor in multimodal editing. 
By directly adopting locality metrics from text-only settings, these methods fail to account for the unique challenges of multimodal inputs, where such simplistic extensions are often inadequate.

{Specifically,  existing  locality evaluation focuses only on whether the final output stays the same, without examining whether the model's inference process or modality usage has changed. 
As shown in Figure~\ref{example}, after we update the output for question-image pair, the model cannot  output  the correct answer ``Black'' instead of ``Green''.}
{Even when presented with a rephrased question \emph{``What color are the shoes of this boy?"} alongside an image of yellow shoes (cf. Figure~\ref{example} (b)), the edited model still outputs ``Black'', disregarding the contradictory visual evidence. 
More critically, when given only the edited text (with no accompanying image), the model still incorrectly responds with ``Black'', indicating that the post-edit model tends to overfit to the edited fact and ignore visual evidence when faced with semantically similar inputs.
This issue is particularly problematic in multimodal models, where both image and text are jointly used for prediction. An edit that targets only the textual representation can lead the model to over-rely on language and disregard visual cues, even when the output appears correct.
Such behavior reflects a breakdown in cross-modal balance, a key aspect of model fidelity in multimodal settings.}

{To address this issue, we propose a novel evaluation framework for analyzing the locality of multimodal model editing (MMED). We introduce \textbf{De-VQA}, a Dynamic Evaluation framework for visual question answering (VQA), which automatically selects adversarial samples that are similar (but not identical) to the editing data in either the text or image modality. To assess a post-edit model's multimodal ability, we define three locality metrics: \emph{Random-Image Locality (RI-Loc)}, \emph{No-Image Locality (NI-Loc)}, and \emph{Consistent-Image Locality (CI-Loc)}. These are quantified using seven distinct data types designed to probe different aspects of locality. 
Using De-VQA, we uncover the phenomenon of \textbf{transient blindness}, a degradation in multimodal locality where edits to a model’s textual knowledge cause it to temporarily ``ignore’’ or under-utilize visual inputs during inference.}
{To investigate the root causes of transient blindness, we employ token attribution tracing and find that current editing methods disproportionately alter textual representations while leaving visual representations largely unaffected. This imbalance in hidden state updates causes edited models to over-rely on textual cues, giving rise to transient blindness. We introduce an adversarial loss that amplifies the influence of visual inputs during editing, thereby reducing this effect. Experiments on two datasets and across two multimodal models demonstrate that our method effectively mitigates transient blindness, improving both the robustness and reliability of the edited models.}

In summary, our main contributions are as follows:

\smallskip \noindent 
{\textbf{(1)} We propose \textbf{De-VQA}, a dynamic evaluation framework designed to detect breakdowns in locality during multimodal model editing (MMED). It introduces three evaluation dimensions
operationalized through seven carefully constructed data types. Together, these components form the first comprehensive benchmark for assessing locality preservation in MMED.}

\smallskip
\noindent { \textbf{(2)} We characterize locality failures as \emph{transient blindness}: a phenomenon in which post-edit models overfit to textual inputs resembling the edit, while neglecting visual information. Our analysis reveals that transient blindness stems from imbalanced updates between the textual and visual modalities during the editing process.}

 \smallskip
\noindent \textbf{(3)} {We propose a locality loss that balances cross-modal updates. Extensive experiments show that our approach consistently mitigates transient blindness and improves locality preservation by 17\% on average across multiple models and datasets, while maintaining edit accuracy.}

\section{Multimodal Model Editing}
\label{met}

\noindent\textbf{Task Definition. }
{Multimodal model editing (MMED) aims to correct the output of a multimodal model $f(\cdot;\theta)$ with parameters $\theta$ without 
full retraining or fine-tuning, while preserving the model’s original outputs for unrelated inputs.
MMED employs an editing function $g(\cdot)$ to achieve $g(f(x_e))=a$, where $x_e$ is the input for the edits and $a$ is the desired  output.
For inputs $x_o$ unrelated to $x_e$, the goal is to maintain their outputs unchanged, i.e. $g(f(x_o))=y =f(x_0)$.
Note: $x_*$  is composed of the  input  image $x^i_*$ and the input text $x^t_*$.}

\smallskip \noindent \textbf{Evaluation Metrics. }
{To evaluate the performance of an editing method $g(\cdot)$,  prior work uses the following datasets: the \emph{editing dataset} $D_e$ composed  of tuples $ \langle x^i_e, x^t_e, y_e, a \rangle $, the \emph{semantically equivalent dataset} $D_g$ composed of tuples $\langle x^i_g, x^t_g, y_g, a \rangle $ and the \emph{locality dataset}   $D_o$ composed of tuples $\langle x^i_o,x_o^t,y_o\rangle$. They use the following metrics:}
{\begin{itemize}[itemsep=0.5ex, leftmargin=5mm]
    \item 
\textbf{Reliability} measures how effectively $ g(\cdot) $ updates the output of model to $ a $ instead of $ y_e $, for each data element in $ D_e $. This is  calculated as:
\begin{equation}
    \textrm{Rel}=\mathbb{E}_{ \langle x^i_e, x^t_e, y_e, a \rangle \in D_e}[\mathbf{1}_{g(f(x^i_e, x^t_e), y_e)=a}].
\end{equation}

\item
\textbf{Generality} assesses the consistency of predictions from the post-edit model $ g(f(\cdot)) $ when presented with semantically equivalent inputs from $ D_g $. 
This includes modified text and images, known as \emph{text-generality (T-Gen)} and \emph{image-generality (I-Gen)}, respectively:
\begin{equation}
    \textrm{T-Gen}=\mathbb{E}_{ \langle x^i_e,  x^t_g, y_e, a \rangle \in D_g}[\mathbf{1}_{g(f(x^i_e, x^t_g),y_e)=a}].
\end{equation} 
\begin{equation}
    \textrm{I-Gen}=\mathbb{E}_{ \langle  x^i_g, x^t_e, y_e,a \rangle \in D_g}[\mathbf{1}_{g(f(x^i_g, x^t_e),y_e)=a}].
\end{equation} 
\item
\textbf{Locality} evaluates how well $ g(f(\cdot)) $ maintains the original predictions of $ f(\cdot) $ for each data in $ D_o $. 
Locality is measured using Text-Locality (T-Loc) and Image-Locality (I-Loc), which are expressed as follows:
\begin{equation}
    \textrm{T-Loc}=\mathbb{E}_{ \langle x^t_o, y_o \rangle \in D_o}[\mathbf{1}_{g(f(x^t_o),y_o)=f(x^t_o)}].
    \label{T-Loc}
\end{equation} 
\begin{equation}
    \textrm{I-Loc}=\mathbb{E}_{ \langle x^i_o, x^t_o,y_o \rangle \in D_o}[\mathbf{1}_{g(f(x^i_o, x^t_o),y_o)=f(x^i_o, x^t_o)}].
    \label{I-Loc}
\end{equation} 
\end{itemize}}
These metrics primarily assess whether the model's outputs remain unchanged on inputs unrelated to the edit, without probing shifts in inference or modality usage. 
However, in multimodal models, we find that edits targeting textual inputs can induce an over-reliance on language, causing the model to overlook visual cues, even when the outputs appear correct.  
Therefore, beyond output preservation, it is crucial to evaluate whether the model continues to appropriately integrate both visual and textual modalities after editing.

\begin{figure}[t]
\centering
  \includegraphics[scale=1.3]{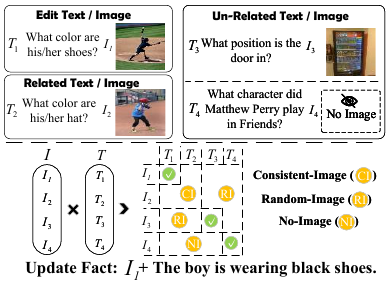} 
  \caption {{Overview of De-VQA: dynamic sampling of  related and unrelated image-text pairs $(T_i,I_j)$,  $i,j\in\{2,3,4\}$, for the edited pair $(T_1I_1)$. 
  Consistent-Image indicates that either the image or textual input  are related to the edited data.  
  Random-Image represents cases of image-text mismatch.  No-Image denotes text-only inputs without any accompanying image.}
  }
  \label{dataset}
\end{figure}


\section{Dynamic Evaluation Framework }
\label{metric}
{To overcome these limitations, we propose three new locality metrics that provide a more rigorous evaluation of modality utilization and the robustness of edited models.  
To support these metrics, we introduce \textbf{De-VQA}, a novel framework that automatically generates test cases tailored to specific model edits.  
The details are as follows.
} 

\smallskip \noindent \textbf{Dynamic Data Sampling.} 
{We construct evaluation samples by selecting image-text pairs with varying 
similarity to   edited data, covering 
related and 
unrelated content in both  modality, to allow a comprehensive locality assessment.}
{For example, in Fig.~\ref{dataset}, suppose the edit is: \emph{The boy is wearing black shoes.} applied to image $I_1$. We label the corresponding image and text as $I_1$ and $T_1$, respectively. Through dynamic sampling, we retrieve a semantically similar text $T_2$, along with irrelevant texts $T_3$ and $T_4$, and group them into a text set: $T = \{T_1, T_2, T_3, T_4\}$. }

{Correspondingly, we obtain images $I_2$, $I_3$, and $I_4$ (the image sources for $T_2$, $T_3$, and $T_4$) forming the image set $I = \{I_1, I_2, I_3, I_4\}$. 
Note that both $T_3$ and $T_4$ are unrelated to the edit, with a key distinction: $T_3$ is paired with an image $I_3$ (capturing  I-Loc), while $T_4$ is a standalone text query without an associated image (capturing  T-Loc). 
We use the retriever from IKE~\cite{ike} to identify similar data pair to the $T_1$ as the related sample $T_2$ and image $I_2$. }

\smallskip \noindent \textbf{Data Construction Evaluation.}
We compute the Cartesian product $\zeta$ of the text set $T$ and image set $I$ to obtain all possible combinations of inputs:
\begin{equation}
T \times I = \zeta = \{(T_i, I_j)\ |\ T_i \in T,\ I_j \in I\},
\end{equation}
where $\times$ denotes the Cartesian product. 
As illustrated in Fig~\ref{dataset}, this results in a total of 16 unique text-image pairs. 
We exclude the pair $(T_1, I_1)$ (Rel) because it is used to evaluate \emph{reliability}. 
Thus, for locality evaluation, we define the set:
\begin{equation}
\textit {Locs} = \zeta \setminus \{(T_1, I_1)\}.
\end{equation}
Among the remaining pairs, existing locality evaluations typically focus only on two cases: $(T_3, I_3)$ (I-Loc) and $(T_4, I_4)$ (T-Loc), which represent fully unrelated inputs in the image and text modalities, respectively.  
While these evaluations verify that the edit does not affect unrelated inputs, they are limited in scope and fail to assess how the edit generalizes to semantically related yet distinct multimodal combinations.

\smallskip \noindent \textbf{Evaluation Metrics.} To provide a more comprehensive assessment of locality in the multimodal setting, we split the the remaining 13 combinations  in $\textit{Locs}$ into three classes based on their relationship to the edits:  \emph{random-image locality, no-image locality}, and \emph{consistent-image locality}.

\smallskip
\noindent
{$\mathsf{Random\text{-}Image \text{ } Locality \, \text{(RI-Loc)}}$ refers to input pairs in which the image and text are mismatched and mutually irrelevant (e.g., $T_1I_3$, $T_2I_3$, $T_3I_1$, $T_4I_1$). 
As shown in the RI region of Fig~\ref{dataset}, this category includes a total of seven such combinations. 
RI-Loc evaluates whether the post-edit model disproportionately relies on the text input when the accompanying image offers no relevant information. 
In De-VQA, rather than exhaustively evaluating all RI combinations, we focus on two representative cases: \emph{the random-image replacement} ($T_1I_3$) and the \emph{random-text replacement} ($T_3I_1$), where either the edited image or text is replaced with random inputs.  
These scenarios serve as effective probes for measuring overreliance on a single modality. We define RI-Loc as:}
\begin{equation*}
    \textrm{RI-Loc}=\mathbb{E}_{ \langle x^i, x^t,a \rangle  \in \{T_1I_3, T_3I_1\}}[\mathbf{1}_{g(f(x^i, x^t),y) \neq a}].
    \label{RI-Loc}
\end{equation*}

\noindent {$\mathsf{No\text{-}Image \text{ } Locality}$ (NI-Loc) 
refers to image-free inputs, text-only queries paired with a missing or null image input. 
This category includes three combinations: $T_1I_4$, $T_2I_4$, and $T_3I_4$. NI-Loc evaluates whether the model can avoid producing incorrect outputs in the absence of visual context, thereby ensuring it does not over-rely on textual cues alone. 
In De-VQA, we focus on two critical scenarios where the image input is entirely removed: \emph{Edited Text without Image} ($T_1I_4$) and \emph{Similar Text without Image} ($T_2I_4$). These cases are designed to assess the model's robustness under image-free conditions and are computed as follows:}
\begin{equation*}
    \textrm{NI-Loc}=\mathbb{E}_{ \langle x^t,y,a \rangle \in  \{T_1I_4,T_2I_4\}}[\mathbf{1}_{g(f(x^t),y) \neq a}].
    \label{NI-Loc}
\end{equation*} 

\noindent {$\mathsf{Consistent\text{-}Image \text{ } Locality} $ (CI-Loc)
captures input pairs where the image, the text, or both are semantically similar to the edited data. This includes three key combinations: $T_1I_2$, $T_2I_1$, and $T_2I_2$. 
To construct CI-Loc samples, we first retrieve semantically similar textual questions ($T_2$) using the IKE retriever. 
Then, we obtain their corresponding paired images ($I_2$) from the dataset.  
CI-Loc assesses whether the edit has unintentionally impaired the model's ability to accurately process visual information when presented with variations of the original textual input. In De-VQA, due to the strong semantic relevance of these samples to the edited input, we evaluate all combinations formed by $T_1$, $T_2$ and $I_1$, $I_2$, expressed as:}
\begin{equation*}
    \textrm{CI-Loc}=\mathbb{E}_{ \langle x^i, x^t,y,a \rangle   \in \{T_1I_2,T_2I_1,T_2I_2\} }[\mathbf{1}_{g(f(x^i_e, x^t_r),y)\neq a}].
    \label{CI-Loc}
\end{equation*} 
{Testing with De-VQA reveals that while existing editing methods excel on original metrics, they perform poorly on the above metrics. 
As shown in Fig \ref{example} (a), edited models exhibit transient blindness, an overfitting to textually similar queries while ignoring accompanying images. }

\begin{figure}[t]
\centering
  \includegraphics[scale=0.9]{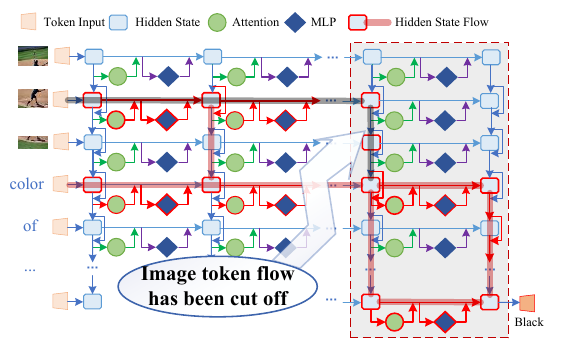} 
  \caption {{Causal information flow in multimodal models.  
The red paths highlight the causal trace originating from image tokens. 
After editing high layer (gray area), the causal influence from image tokens is blocked (indicated by the black paths),  
while the flow from text tokens remains unaffected. }
    }
  \label{flow}
\end{figure}

\section{Alleviating  Transient Blindness in MMED}

{Through our evaluation framework, we found the existence of transient blindness in post-edit multimodal models. 
This section analyzes its causes and proposes mitigation strategies. 
We first examine the relative influence of text and image tokens on model outputs both before and after editing. 
Our analysis reveals that post-edit models exhibit an increased reliance on textual information, consequently diminishing the impact of visual inputs. 
Based on this observation, we propose an adversarial loss to balance the model's attention to textual and visual knowledge after editing.} 

\subsection{Token Attribution in Multimodal Models}
\label{attribution}
{To assess modality contributions in Large Multimodal Models (LMMs), we trace token-level influence on the output by analyzing which tokens most affect the final hidden state.
Given input tokens $x=\{t_0^i,\dots,t_m^i, t_0^w,\dots,t_n^w\}$, where $t_j^i, \,j \in [0,m]$ are image tokens and $t_k^w, \, k \in [0,n]$  are text tokens, the hidden state at layer $\ell$ is:}
\begin{equation*}
h^\ell = \text{MLP}(h^{\ell-1}) + \text{Attn}(h^{\ell-1}) + h^{\ell-1} = m^\ell + a^\ell + h^{\ell-1}.
\label{trans}
\end{equation*}
{We focus on the output token $h^L_N$  ($N=m+n$) and backtrack important contributors using a queue-based token tracing. For each token $h_i^\ell$ in the queue $Q$, we extract components via:}
$
{Hook}(h_i^\ell) = \{h^{\ell-1}_i, m^\ell_i, a^\ell_i\},
\label{hook}
$
and compute each component’s contribution score:
\begin{equation}
{Distance}(h_i^\ell, a) = \frac{L_2(h_i^\ell - a)}{\sum_{j \in \{h, m, a\}} L_2(h_i^\ell - j)} + \cos \langle h_i^\ell, a \rangle,
\label{scores}
\end{equation}
which combines geometric distance and representational alignment.
{Tokens with high scores are recursively added to $Q$, forming a critical influence path. We compute the image-to-text ratio among influential tokens across layers. We find editing reduces the contribution of image tokens, shifting the model's reliance toward textual input (see Figure~\ref{flow}).}

\subsection{Adversarial Enhancement for Mitigating Transient Blindness}
\label{smend}

{Token attribution analysis reveals that post-editing, the model over-relies on textual input while neglecting visual cues. 
To restore cross-modal balance, we propose an adversarial sample augmentation strategy inspired by recent multimodal regularization methods \citep{pi2025strengthening,chen2024we,wu2024noiseboost}.}
{We build on MEND \citep{mend}, a hypernetwork-based editing method (Fig.~\ref{baseline}), which computes low-rank parameter updates from the edit loss $\mathcal{L}_e = -\log p_{\theta'}(y_e \mid x_e^i, x_e^t)$. To preserve unrelated behaviors, MEND introduces a locality constraint: $\mathcal{L}_{loc}=\textbf{KL}(p_{\theta}(\cdot\mid x_o^i,x_o^t)\mid\mid p_{\theta'}(\cdot\mid x_o^i,x_o^t)) +\textbf{KL}(p_{\theta}(\cdot\mid x_o^t)\mid\mid p_{\theta'}(\cdot\mid x_o^t))$.}
{To balance the contribution between image and text modalities in the locality constraint, we use a KL divergence loss to ensure that the model’s output distribution remains consistent before and after editing:
\begin{equation}
\mathcal{L}_{loc}^M = \textbf{KL}\left(p_{\theta}(\cdot \mid x) || p_{\theta'}(\cdot  \mid x)\right),
\end{equation}
where $\theta$ and $\theta'$ are the model parameters before and after editing, and $x=(x_*^t,x_*^i)$ denotes a multimodal input composed of the edited textual question $x_*^t$ and an unaltered image $x_*^i$. }
{To provide diverse locality constraints, we select representative samples of $x$ from three different types:
(i) RI, e.g., $T_1I_3$,
(ii) CI, e.g., $T_2I_2$ and
(iii) NI, e.g., $T_1I_4$.
We combine the KL losses computed from these three types of inputs as the final locality constraint $\mathcal{L}_{loc}^M$. 
Analysis of combinations is provided in Section~\ref{alb_sec}.
The final objective is:
\begin{equation}
\textrm{Loss} = \lambda_1\mathcal{L}_e + \lambda_2\mathcal{L}_{loc} + \lambda_3\mathcal{L}_{loc}^M,
\label{loss_fun}
\end{equation}}
where we set $\lambda_1$ to 0.1 and $\lambda_2,\lambda_3$ to 1. This regularization encourages the model to rely more on image features when text is ambiguous or mismatched, mitigating transient blindness and reinforcing visual grounding after edits.

\begin{figure*}[htbp]
\centering
  \includegraphics[scale=0.3]{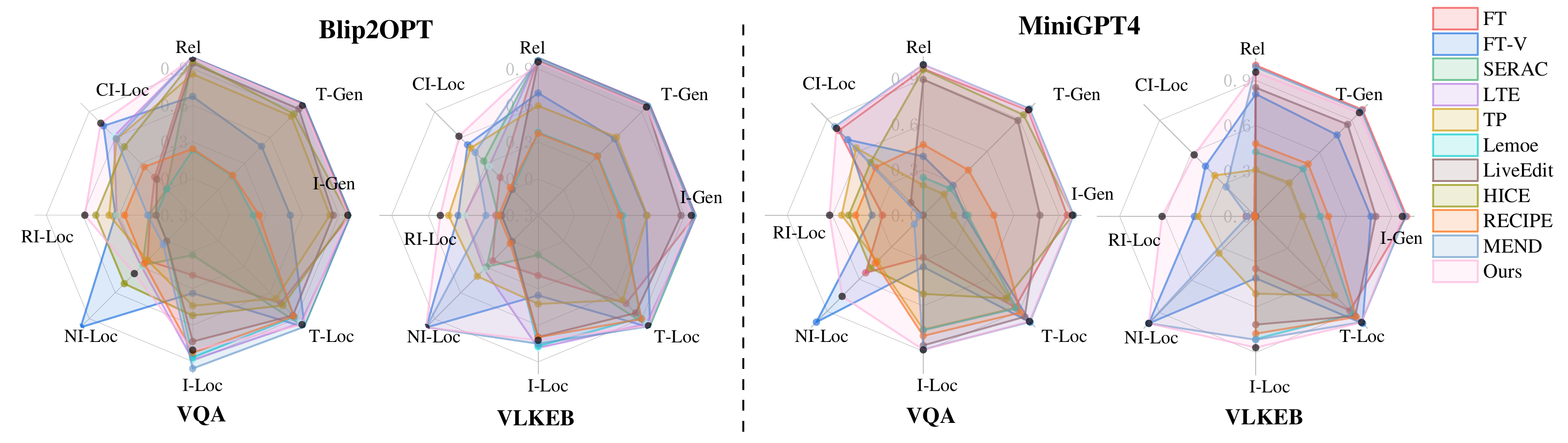} 
  \caption {Main experiment results on Blip2OPT and MiniGPT4. 
 {The lower performance of existing editing methods on \{CI ($T_2I_2$), RI ($T_1I_3$), NI ($T_1I_4$)\}-Loc compared to the better performance on \{T,I\}-Loc reflects the inadequacies of the original locality evaluation.
  Our method (black node) can achieve more comprehensive performance in terms of locality.} 
  }
  \label{Main_results}
\end{figure*}

\section{Experiments}
{To evaluate the effectiveness of De-VQA,  we aim to explore the following research questions:
\begin{itemize}[itemsep=0.5ex, leftmargin=5mm]
\item
\textbf{RQ1:}  What limitations of existing locality evaluation methods can De-VQA reveal?
\item
\textbf{RQ2:} How does our method perform under the De-VQA framework compared to prior editing approaches?
\item
\textbf{RQ3:} Why does model editing lead to transient blindness, and how does our method mitigate this issue?
\end{itemize}}

\subsection{Experimental Setup}
\noindent \textbf{Dataset. } 
{To reassess existing editing methods and quantify the transient blindness induced by model editing. }
{We apply  De-VQA based on   VQA \citep{MMED}, which contains 6,346 training data entries and 2,093 test data entries, and  VLKEB \citep{huang2024vlkeb}, which includes 5,000 training data  and 3,174 test data. }
And the VQA is prioritized as it need strict cross-modal collaboration, but captioning/OCR rely less on balanced text-image use.

\smallskip
\noindent \textbf{Baselines.}  {Our evaluation includes two multimodal models: Blip2OPT \citep{blip2} and MiniGPT4 \citep{minigpt4} and Qwen-2.5-VL \citep{Qwen-VL} which are popular models in multimodal model editing settings. }
{We use \texttt{all-MiniLM-L6-v2}~\citep{reimers2019sentence} to select the evaluation samples based on cosine similarity.}
{We use the editing methods FT \citep{FT}, MEND \citep{mend}, TP \citep{patch}, LTE \citep{jiang-etal-2024-learning}, RECIPE \citep{chen-etal-2024-lifelong},  Lemoe \citep{wang-li-2024-lemoe}, LiveEdit \citep{Chen_2025_CVPR}, HICE \citep{ma2025comprehendedit} and  SERAC \citep{serac}.}  {We edit a single instance at a time.}
{All experiments were conducted on a single NVIDIA A100 (40GB) GPU.
Hyperparameters and implementation details are kept consistent with these baselines to ensure fair comparison}\footnote{DeVQA is a plug-and-play framework, so we evaluate each methods using their original settings under a single-edit setup.
}.
Additional implementation details are provided in the Appendix \ref{Baselines}.

\subsection{Main Results}
To address \textbf{RQ1}, we conduct experiments on the VQA and VLKEB. 
Main results demonstrate that {\textit{De-VQA uncovers critical limitations of existing evaluation metrics, and our method outperforms baselines in mitigating transient blindness while maintaining edit accuracy.} As shown in Figure \ref{Main_results}, while most editing methods demonstrate promising performance on traditional metrics such as Rel, T-Gen, and I-Gen, these metrics fail to fully capture the nuanced impacts of model editing on locality. 
The limitations become evident when we evaluate methods using De-VQA, which identifies transient blindness.} 

{For instance, MEND and SERAC achieves high scores of 0.99 on most metrics. 
Similarly, LTE and Lemoe maintain relatively balanced performance across these metrics, with scores above 0.9. 
However, these methods generally show poor locality preservation under De-VQA. 
For example, most methods only get a score lower than 0.5, SERAC, LiveEdit, Lemoe obtain very low scores on NI-Loc ($<$ 0.3) , RI-Loc ($<$ 0.1), and CI-Loc ($<$ 0.2), indicating that they fail to consistently preserve unrelated knowledge in varying contexts. 
This exposes a significant limitation in existing evaluation protocols, which DeVQA explicitly uncovers.}
{NI-Loc for FT-V is special, because updating the visual module alone does not affect the model's output when given only text inputs, resulting in a perfect NI-Loc score of 1.}  

{\textit{Our approach demonstrates superior performance in De-VQA metrics, effectively alleviating the problem of transient blindness.} 
Unlike existing methods, our strategy balances the contributions of both textual and visual information during editing,}
{improving locality robustness with scores of 0.7 on NI-Loc and CI-Loc, 0.6 on RI-Loc.
As a result, our method maintains high editing success rates and preserves the model's multimodal capabilities, as evidenced by consistently strong performance across all De-VQA metrics.}

\subsection{Analyzing Locality with De-VQA}
\begin{figure}[t]
\centering
  \includegraphics[scale=0.2]{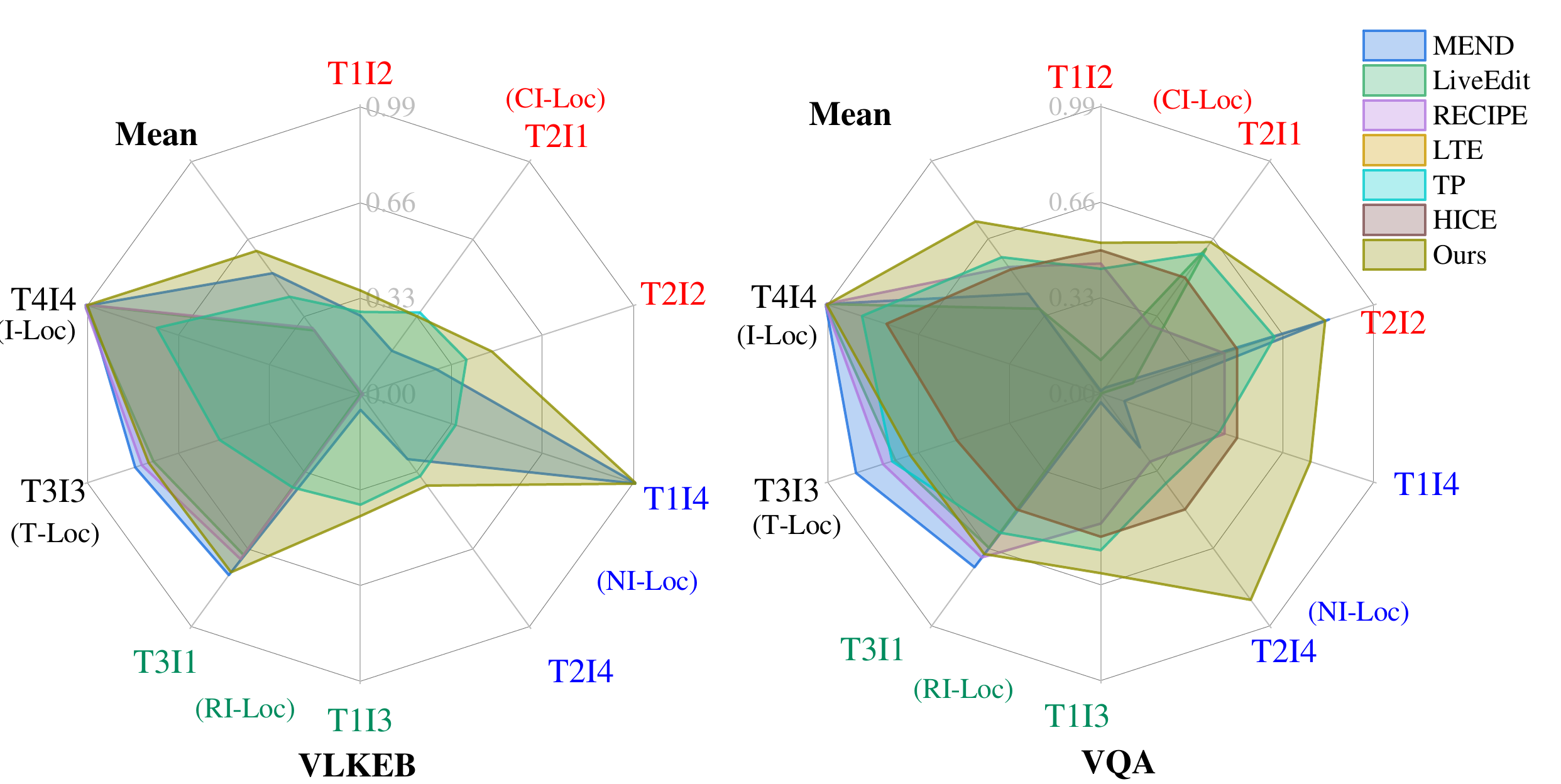} 
  \caption { 
  {Locality metric performance comparison   on MiniGPT4. 
Different colors denote metric types.}
  }
  \label{devqa_details}
\end{figure}

To address \textbf{RQ2}, We conduct a comprehensive evaluation of different locality metrics, as summarized in Figure~\ref{devqa_details}. 

{MEND, RECIPE and LiveEdit exhibit poor performance across several metrics (e.g., $T_2I_1$, $T_1I_3$), with scores falling below 0.5, and consistently underperform on CI-Loc with a score close to 0.3. 
This indicates severe overfitting, as these methods consistently produce the edited answer even when given inputs only marginally similar to the original edited prompt.} 
{TP achieves strong performance across all metrics, obtained a mean of around 0.5 on all metrics. 
We attribute this to its strategy of updating only a small subset of neurons for each individual data point, resulting in minimal perturbation to the model parameters. 
As a result, it is better able to preserve the original capabilities of the model compared to other methods.
HICE extends ICE \citep{ike} by introducing a binary classifier to determine whether contextual information should be incorporated. 
However, we observe that its prediction scores are often centered around 0.5, suggesting that the model fails to truly understand the multimodal content.
Compared to editing methods such as MEND, RECIPE, and LTE, our method achieves higher CI-Loc scores ($T_1I_2$ and $T_2I_2$), achieve average 15\% improvement, indicating improved cross-modal consistency. 
However, performance on $T_2I_1$ remains limited, which we attribute to the difficulty of fine-grained entity understanding when queries involve visually similar entities.}

{Overall, our method outperforms all baselines by addressing the \textbf{multimodal inconsistency problem}. By incorporating cross-modal locality modeling and consistency-aware loss functions, it avoids overfitting to the edited prompt and maintains both visual and textual fidelity. This makes it more robust and reliable across a diverse set of locality scenarios.
See Appendix \ref{typesres} for more results and discussions.}

\begin{figure*}[t]
\centering
  \includegraphics[scale=0.3]{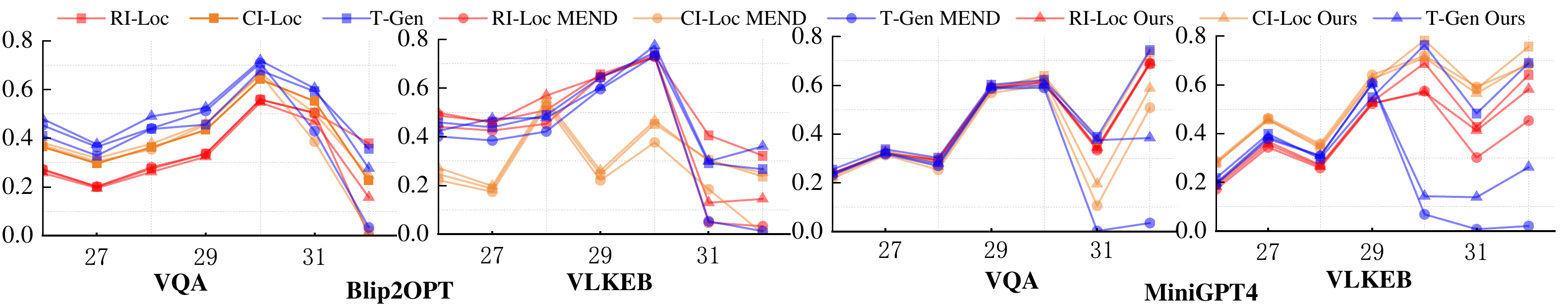} 
  \caption { 
    Contribution of image tokens in Blip2OPT and MiniGPT4, with evaluation across RI-Loc, CI-Loc, and T-Gen metrics. 
    Squares, circles, and triangles denote results from the original model, MEND, and our method, respectively.
  }
  \label{token_attributions}
\end{figure*}

\subsection{Analysis of Transient Blindness}
To address \textbf{RQ3}, we employ token contribution analysis to measure the impact of each modality on model's output before and after editing.

{We first verify the effectiveness of our token attribution method introduced in Section~\ref{attribution}. 
By masking non-critical tokens in specific layers and measuring the resulting performance drop (Table~\ref{tokens_mask}), we observe that masking only the top four layers retains approximately 87\% of the original performance, while masking more layers leads to a gradual decline. 
This indicates that the identified tokens in higher layers are indeed critical to the model’s predictions and that our attribution method accurately captures influential inputs at the editing-relevant layers. }

{Building on this, we further investigate these critical tokens by analyzing attribution scores for visual and textual tokens (particularly in the upper layers) measured by the ratio of image to text token attributions.
As shown in Figure~\ref{token_attributions}, both our method and the original (pre-edit) model maintain strong contributions from image tokens in layers beyond 29, evidenced by higher scores for visual tokens (squares and triangles) compared to textual ones (circles). 
In contrast, MEND significantly reduces the contribution of visual tokens post-editing, leading to a text-dominated output, and make the transient blindness.} 
More results across all layers are provided in Appendix \ref{tbs}.

\begin{table}[t]
\resizebox{.99\columnwidth}{!}{
\begin{tabular}{cccccccc}
\toprule
Layer & 29-32  & 25-32  & 20-32  & 15-32  & 10-32  & 5-32   & 0-32   \\ \midrule
Model & \multicolumn{7}{c}{Blip2OPT}                                 \\ \midrule
VQA   & 0.9506 & 0.5629 & 0.2382 & 0.2079 & 0.1773 & 0.0169 & 0.0017 \\
VLKEB & 0.8704 & 0.8065 & 0.6517 & 0.5218 & 0.3559 & 0.0473 & 0.0060 \\ \midrule
Model & \multicolumn{7}{c}{MiniGPT4}                                 \\ \midrule
VQA  & 0.8966 & 0.8733 & 0.8233 & 0.7353 & 0.6453 & 0.5927 & 0.5820 \\
VLKEB & 0.8866 & 0.7740 & 0.6920 & 0.4707 & 0.2913 & 0.2267 & 0.1467 \\ \bottomrule
\end{tabular}}
\caption{{Performance Change After Masking Unimportant Tokens. 
The results indicate that utilizing only the tokens we have pinpointed for prediction can maintain over 87\% of the performance at higher layers.}}
\label{tokens_mask}
\end{table}

\begin{figure*}[t]
\centering
  \includegraphics[scale=0.4]{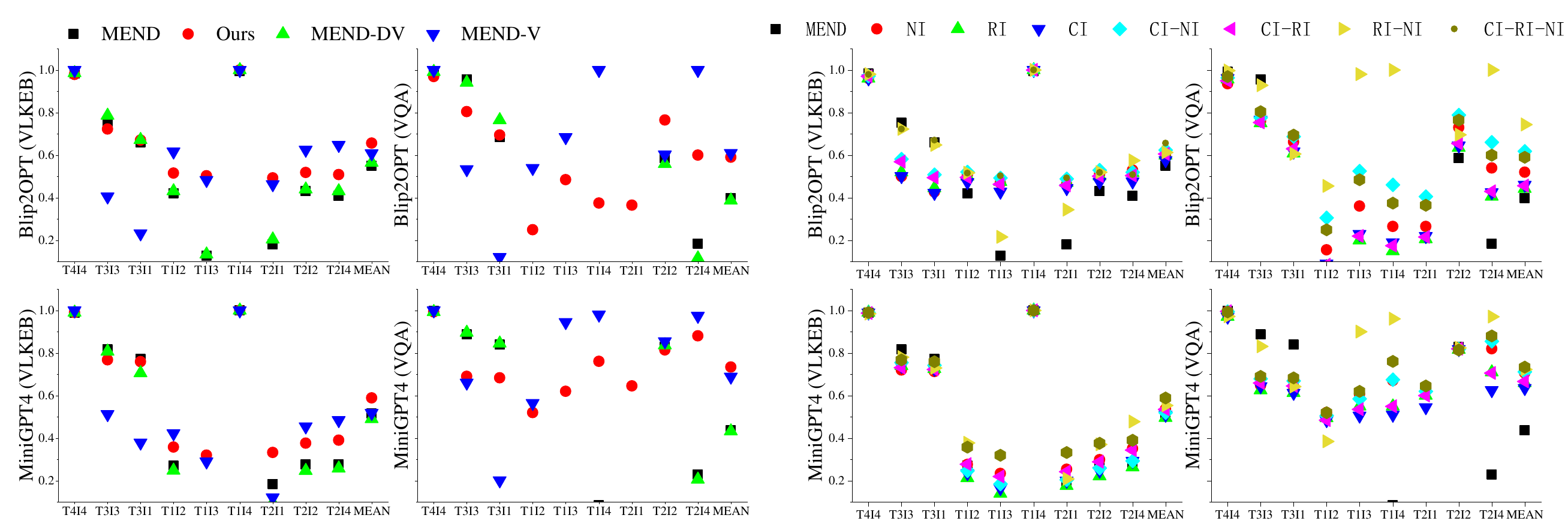} 
  \caption { 
  Ablation study results: (left) module updating comparison, where D refers to the last three layers of the LLM, V refers to the last three layers of the visual encoder, and DV is their combination. (right) loss function comparison. 
  }
  \label{ablss}
\end{figure*}

\subsection{Ablation Study}
\label{alb_sec}
\smallskip\noindent \textbf{Editing Different Modules. }
{We independently update the visual encoder (V), text encoder (T), and their combination (DV) using MEND, following the same setup as \citet{MMED}. 
As shown in Figure~\ref{ablss}, updating V or DV significantly improves locality metrics, indicating the importance of aligning multimodal components. 
However, such updates lead to a drop of over 20\% in relevance and generality, suggesting degraded model consistency. 
In contrast, editing LLM only with appropriate loss constraints yields a better balance between editing effectiveness and model retention. 
Note that when only the image encoder is updated, $T_1I_4$ and $T_2I_4$ reach 1 due to unchanged LLM parameters.
}

\smallskip\noindent \textbf{Effect of Loss Functions. }
{We investigate the impact of incorporating RI, NI, and CI losses individually and in combination (cf. Figure~\ref{ablss}). 
The training data we use for each type of locality is: RI ($T_1I_3$), CI ($T_2I_2$), and NI ($T_1I_4$). 
Incorporating all three losses yields the best and most stable performance, achieving the highest scores in three out of four settings and ranking third on the Blip2OPT model with VQA. 
While using only RI+NI produces comparable results, it introduces instability on metrics such as $T_1I_2$ and $T_1I_3$, and performs well only on MiniGPT-4 with VLKEB. 
This instability arises from challenges in distinguishing edits involving the same image but differing attributions, for instance, identifying `black shoes' versus `yellow shoes' with the edits \emph{The boy is wearing black shoes}. 
In such cases, more fine-grained constraints are necessary, which are effectively addressed by incorporating the CI loss. 
These results underscore the importance of jointly optimizing across diverse data types. 
By combining RI, NI, and CI losses, we enforce stronger consistency and robustness in editing performance across multimodal contexts. Thus we employ the CI-RI-NI as our loss combination.}

\smallskip\noindent \textbf{Retrieval Consistency Analysis} 
We also evaluate the consistency between text-based and image-based retrieval in multimodal model editing. 
We found that the average similarity between samples retrieved via image-based similarity and those retrieved via text-based encoding, as computed by CLIP, is 0.89. 
This result demonstrates a high degree of semantic consistency between the two retrieval methods, confirming that both text-based and image-based retrieval can yield comparably relevant samples.

\section{Related Work}
\noindent \textbf{Model Editing. }
\label{dis}
{Model Editing has emerged as a viable strategy for precisely updating LLMs without the expensive resources \citep{wang2023knowledge, zhang2024comprehensive}. 
A key challenge in model editing is how to update information without affecting unrelated data and without compromising the model's performance. \citep{gupta2024model}.
\citet{MMED} extend text-based editing methods to multimodal model editing and preliminarily verify its feasibility. 
However, the edited model only retains the predictions of the pre-edited model for randomly selected data. 
Compared to evaluating the success rate of updates, assessing the locality of editing models is more important for developing robust model editing methods. 
In this paper, we reveal the transient blindness in MMED,
as current evaluations overlook changes in multimodal abilities post-editing, resulting in inaccurate predictions for inputs that are related but differ from the edits.
highlighting the need to develop specific methods for MMED.
}

\smallskip\noindent \textbf{Editing in Multimodal Language Models.}
The development of large language models (LLMs) has spurred notable advances in multimodal LLMs (MLLMs) \citep{xu2023MultiModal,qi2025robust}. 
These models typically leverage an LLM decoder to interpret fused image-text inputs, making the preservation of multimodal capabilities a central concern during model editing. 
Promising future solutions to this challenge include retrieval-augmented generation \citep{yan2025atomic,yan2025prompting} and memory-based mechanisms \citep{ma-etal-2025-memorization-understanding, wang2024large}.

\section{Discussion \& Conclusion}
Our research uncovers 
critical limitations in current MMED evaluations, 
highlighting that existing metrics overlook cross-modal balance degradation and fail to detect transient blindness, a phenomenon where edited models over-rely on textual cues and disregard visual information.
To address this, we propose \textbf{De-VQA}, a plug-and-play dynamic evaluation framework that introduces three comprehensive locality metrics and seven data types to assess multimodal editing effects. Through token attribution analysis, we trace transient blindness to imbalanced updates between textual and visual modalities, and our proposed locality-aware adversarial loss effectively balances cross-modal contributions. Experimental results demonstrate that our method outperforms existing baselines, improving locality preservation by 17\% on average while maintaining high edit accuracy. 

{
Our work not only revolutionises the evaluation of multimodal model editing by exposing transient blindness and refining locality assessment, but also offers a foundation for robust solutions to balance cross-modal updates, enabling more reliable knowledge correction in real-world vision-language applications,
such as correcting misunderstandings in social media image-text posts (e.g., revising a satirical caption to prevent misinformation) and personalized content generation (e.g., learning a new corporate logo and its meaning for consistent marketing creation).
}

\section{Acknowledgments}
This work has been supported by the 
National Natural Science Foundation of China (No.62376144) 
and 
the Natural Language Processing Innovation Team (Sanjin Talents) Project of Shanxi Province, 
the Science and Technology Cooperation and Exchange Special Project of Shanxi Province (No.202204041101016), 
the Key Research and Development Program of Shanxi Province (No.202102020101008), 
National Natural Science Foundation of China (No.62302297).

\bibliography{edit, ins, LMs, other}

\bigskip

\clearpage

\appendix

\section{Locality in MMED}

Model editing has garnered considerable interest in recent years, owing to its capability to efficiently rectify erroneous knowledge encoded in models. \citep{mend, divide, RASE, Insed, MMED}
However, existing work has predominantly focused on methodologies for implementing knowledge updates, while paying insufficient attention to the consequent effects on the model's intrinsic capabilities. \citep{hoelscher2023detecting}
This gap constitutes a central focus of our analysis and underscores a critical direction for future investigation.

At present, model editing techniques—whether applied to unimodal or multimodal settings—remain largely confined to academic exploration. 
In real-world deployments, alternative strategies such as retrieval-augmented \cite{yan2025atomic,yan2025prompting} generation and memory-based mechanisms \cite{ma-etal-2025-memorization-understanding, wang2024large} are often deemed more trustworthy than direct model editing. 
We argue that this discrepancy stems primarily from the inadequate assessment of locality in edited models. 
Thus, our core contribution in this paper lies in innovation for evaluation: identifying critical gaps in existing locality metrics, formalizing transient blindness, and providing an actionable mitigation, all validated by rigorous analysis.

\section{Implementation Details.}
\subsection{Baseline Methods}
\label{Baselines}
\paragraph{FT}
Fine-tuning (FT) \citep{FT} is the most straightforward approach to model editing, involving directly updating the model parameters using newly provided knowledge. FT often results in the unintended alteration of unrelated model behaviors due to its global parameter changes.

\paragraph{SERAC} \citep{serac}
SERAC introduces a hybrid editing framework that avoids modifying the original model parameters. It stores edits in an external memory and uses a classifier to determine whether to apply the stored counterfactuals at inference time. By combining retrieval-based mechanisms with an in-context correction model, SERAC achieves strong performance on QA and factual consistency tasks.

\paragraph{MEND} \citep{mend}
MEND learns a parametric transformation that maps the gradient of a desired edit to a small change in the model parameters. This edit network is trained to generalize across a wide range of edits, allowing efficient, localized updates to model behavior.

\paragraph{T-Patcher} \citep{patch}
T-Patcher is a patch-based editing approach that introduces dedicated "neurons" for each edit, which are inserted into the last feedforward layer of a transformer model. These patch neurons are only activated by specific input patterns corresponding to the edit, enabling strong locality and edit-specific behavior without affecting the original model's parameters.

\paragraph{LTE} \citep{jiang-etal-2024-learning}
Learning to Edit (LTE) proposes a two-stage framework for aligning LLMs with desired edits. In the alignment stage, the model is fine-tuned on parallel data to acquire capabilities in handling in-scope edits, preserving out-of-scope content, and maintaining linguistic fluency. During inference, LTE uses a retrieval mechanism to dynamically inject edit descriptors into the prompt without modifying model parameters.

\paragraph{RECIPE} \citep{chen-etal-2024-lifelong}
RECIPE is a non-parametric, retrieval-based editing method that stores facts or edits in an external memory and re-injects them at inference time via a fusion module. Unlike parameter-editing methods, RECIPE enables lifelong editing without risk of forgetting past edits or corrupting unedited model knowledge. 

\paragraph{LEMoE} \citep{wang-li-2024-lemoe}
LEMoE introduces a MoE-based edit module where each edit is handled by a dedicated expert network. A routing module selects the appropriate expert based on the edit context. LEMoE includes anchor-key-value (KV) memory routing and clustering-based expert selection strategies. 

\paragraph{LiveEDIT} \citep{Chen_2025_CVPR}
LiveEDIT is a lifelong editing framework that combines contrastive routing, soft MoE integration, and local editing loss to support continuous, non-destructive knowledge editing. Unlike earlier MoE methods, LiveEDIT introduces explicit constraints for locality and edit generalization, achieving high reliability and negligible degradation across long edit sequences.

\paragraph{HICE} \citep{ma2025comprehendedit}
HICE is the baseline method introduced in the ComprehendEdit benchmark for multimodal knowledge editing. HICE adopts a two-stage hierarchical prompt injection strategy that embeds edit-related information into the model's input context without altering model weights.

\subsection{Dataset}

De-VQA is a plug-and-play evaluation framework. 
In this work, we utilize EVQA \citep{MMED} and VLKEB \citep{huang2024vlkeb} to assess the locality of multimodal editing. 
When using De-VQA, we dynamically sample and compute only the data related to the edit by selecting the top-1 text-image pair based on similarity between the question text and the edit text. 
These selected pairs are then combined via Cartesian product to obtain a comprehensive dataset for locality evaluation. 
In our experiments, we edit only one data item at a time, and for each type of locality, we sample one instance.

For the selecting, we rely solely on the IKE \citep{ike} method's text similarity ranking to retrieve the most relevant image-text pair for each edit, based on the similarity between the edit text and question text. 
Importantly, we always select the top-1 pair until the retrieved image-text pair's answer is different from the edited answer, ensuring the data is relevant yet challenging for locality evaluation.

\subsection{Hyperparameters}

We update the last three layers of the decoding module in the multimodal model (Blip2OPT and MiniGPT4), which we denote as `-D'. (Refer to Table \ref{d4},\ref{db})
We follow the settings and parameters of EasyEdit \footnote{\url{https://github.com/zjunlp/EasyEdit}}. 
Additionally, FT also updates the parameters of the last three layers of the image encoding module, denoted as `-V'. 
All experiments were conducted on a single NVIDIA A100 (40GB) GPU.
All hyperparameters and implementation details are kept consistent with these baselines to ensure fair comparison. 
Please refer to LiveEdit \footnote{\url{https://github.com/qizhou000/LiveEdit/tree/main}} and \footnote{ComprehendEdit\url{https://github.com/yaohui120/ComprehendEdit}}

The Blip2OPT \citep{blip2} and MiniGPT-4 \citep{minigpt4}  consist of a pre-trained Image Encoder, a pre-trained Large Language Model, and a learnable Q-Former\citep{q-former}. 
During training, both models freeze the vision and LLM parameters and train either the Q-Former or the projection layer.

\begin{table}[ht]
\centering
\resizebox{.9\columnwidth}{!}{

\begin{tabular}{@{}cl@{}}
\toprule
Category & Parameters \\ \midrule
D        & \texttt{opt\_model.model.decoder.layers.29.fc1.weight} \\
         & \texttt{opt\_model.model.decoder.layers.29.fc2.weight} \\
         & \texttt{opt\_model.model.decoder.layers.30.fc1.weight} \\
         & \texttt{opt\_model.model.decoder.layers.30.fc2.weight} \\
         & \texttt{opt\_model.model.decoder.layers.31.fc1.weight} \\
         & \texttt{opt\_model.model.decoder.layers.31.fc2.weight} \\
\addlinespace
V        & \texttt{Qformer.bert.encoder.layer.9.output\_query.dense.weight} \\
         & \texttt{Qformer.bert.encoder.layer.10.output\_query.dense.weight} \\
         & \texttt{Qformer.bert.encoder.layer.11.output\_query.dense.weight} \\
\bottomrule
\end{tabular}}

\caption{List of Edited Parameters for Blip2OPT}
\label{db}
\end{table}

\begin{table}[ht]
\centering
\resizebox{.9\columnwidth}{!}{
\begin{tabular}{@{}cl@{}}
\toprule
Category & Parameters \\ \midrule
D               & \texttt{llama\_model.model.layers.29.mlp.down\_proj.weight} \\
         & \texttt{llama\_model.model.layers.29.mlp.up\_proj.weight} \\
         & \texttt{llama\_model.model.layers.30.mlp.down\_proj.weight} \\
         & \texttt{llama\_model.model.layers.30.mlp.up\_proj.weight} \\
         & \texttt{llama\_model.model.layers.31.mlp.down\_proj.weight} \\
         & \texttt{llama\_model.model.layers.31.mlp.up\_proj.weight} \\
\addlinespace
V        & \texttt{Qformer.bert.encoder.layer.9.output\_query.dense.weight} \\
         & \texttt{Qformer.bert.encoder.layer.10.output\_query.dense.weight} \\
         & \texttt{Qformer.bert.encoder.layer.11.output\_query.dense.weight} \\
\bottomrule
\end{tabular}}
\caption{List of Edited Parameters for MiniGPT4}
\label{d4}
\end{table}

\begin{table}[ht]
\centering
\resizebox{.9\columnwidth}{!}{
\begin{tabular}{cl}
\toprule
Category & Parameters \\
\midrule
D & \texttt{transformer.h.31.mlp.w1.weight} \\
  & \texttt{transformer.h.31.mlp.w2.weight} \\
  & \texttt{transformer.h.30.mlp.w1.weight} \\
  & \texttt{transformer.h.30.mlp.w2.weight} \\
  & \texttt{transformer.h.29.mlp.w2.weight} \\
  & \texttt{transformer.h.29.mlp.w1.weight} \\
\addlinespace 
V & \texttt{transformer.visual.transformer.resblocks.47.mlp.c\_proj.weight} \\
  & \texttt{transformer.visual.transformer.resblocks.47.mlp.c\_fc.weight} \\
  & \texttt{transformer.visual.transformer.resblocks.46.mlp.c\_proj.weight} \\
  & \texttt{transformer.visual.transformer.resblocks.46.mlp.c\_fc.weight} \\
\bottomrule
\end{tabular}}
\caption{List of Edited Parameters for Qwen-VL}
\label{dq}
\end{table}

\section{Alleviating  Transient blindness in MMED}
Through our evaluation framework, we have confirmed the existence of transient blindness in post-edit multimodal models. This section analyzes its causes and proposes mitigation strategies. 
First, we examine the relative influence of text and image tokens on model outputs both before and after editing. Our analysis reveals that post-edit models exhibit an increased reliance on textual information, consequently diminishing the impact of visual inputs. 
Based on this observation, we proposed an adversarial loss to balance the model's attention to textual and visual knowledge after editing. The details are as follows.

\begin{figure}[t]
\centering
  \includegraphics[scale=0.75]{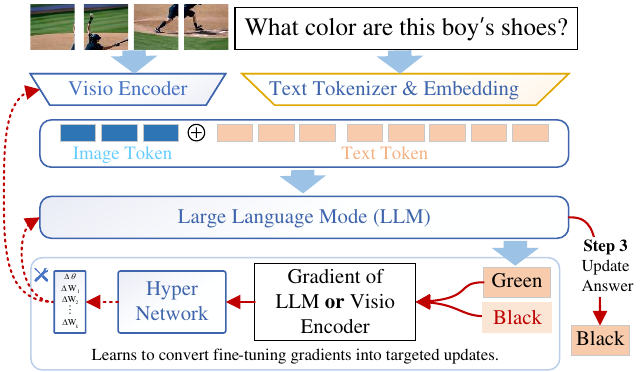} 
  \caption {{HyperNetwork-based Multimodal model editing method. Dashed lines indicate optional update modules.}  }
  \label{baseline}
\end{figure}

\subsection{Token attribution in multimodal models}
\label{attribution}
We aimed to evaluate the contributions of image and text modalities to the model's output by identifying the important tokens that influence the final result and analyzing the proportion of image tokens to text tokens among these key tokens.

Typically, the transformer-based LMM with $L$ layers receives a token sequence $x=\{t_0^i,t_1,^i,..,t_m^i,t_0^w,t_1^w,..,t_n^w\}$ with $m$ image token and $n$ text tokens, we denote $N=m+n$.
The hidden state $h^l$ of layer $l$ is calculate as:
\begin{equation}
h^l = \text{MLP}(h^{l-1}) + \text{Attn}(h^{l-1}) + h^{l-1} = m^l+a^l+h^{l-1},
 \label{trans}
\end{equation}
where $m_l$ and $a_l$ represent MLP and attention outputs, respectively. 
The final prediction $y$ is derived from the last token's hidden state $h^L_{N}$ in the layer $L$ of the $N$-th token.
Thus, we could roughly identify the key tokens within the model by analyzing which tokens have a greater impact on the token $h^L_{N}$ by the following step.

We first initializes a queue $Q$ with the output token $h^L_{N}$.
Then we denote the first element in the queue as $h_i^l$, and extract each type of hidden state for $h_i^l$ using a hook function: 
\begin{equation}
    \textsc{Hook}(h_i^l)=h^{l-1}_i,m^l_i,a^l_i.
    \label{hook}
\end{equation}
Furthermore, the contribution degree of token $h_i^l$ at each layer is calculated as:
\begin{equation}
    \textsc{Distance}(h_i^l,a)= \frac{L_2(h_i^l-a)}{ \sum_{i \in {\{h,m,a\}}}^{}{L_2(h_i^l-i)}}+cos\langle h_i^l, a \rangle,
    \label{scores}
\end{equation}
where $L_2(\cdot)$ is the $L_2$ norm and $cos\langle \cdot \rangle$ is the cosine similarity.
In this way, it could measures both geometric proximity and representational alignment of attention outputs with respect to the token hidden state.
If the value of $\textsc{Distance}(h_i^l,a)$ meets the criteria, we could view other tokens may have great impect the current token and add the tokens with the highest scores (TopK scores) to the queue $Q$. 
These newly added tokens serve as the initial key tokens for the next layer. 
By iterating this process, we approximate a critical path that significantly affects the output. 
The algorithm \ref{alg} gives the details for each layer.

By analyzing this flow and score, we assess the contribution of image and text tokens from each layer to the final output. 
Our analysis revealed that after editing, the contribution of the image tokens is reduced or eliminated, leading the model to rely predominantly on text information to generate answers. 
Figure \ref{flow} gives an illustration.

\renewcommand{\algorithmicrequire}{\textbf{Input:}}
\renewcommand{\algorithmicensure}{\textbf{Output:}}
\begin{algorithm}[t]
\caption{Key Tokens Extraction for Each Layer}
\begin{algorithmic}[1]
\REQUIRE Initial Key Tokens $Q$, Attention Scores $A$, Threshold $\gamma$
\ENSURE Key Tokens $Q'$, Token Scores $S$
\STATE $Q' \leftarrow Q$, $S \leftarrow \emptyset$
\STATE $i \leftarrow 0$
\WHILE{$i < |Q'|$}
    \STATE $t \leftarrow Q'[i]$
    \STATE $(a, m, h) \leftarrow \textsc{Hook}(t)$
    \IF{$\textsc{Distance}(a,t) \geq \gamma$}
        \STATE $T \leftarrow \textsc{TopK}(A_t, t)$
        \STATE $S \leftarrow S \cup \{\textsc{Distance}(a,t)\}$
        \STATE $Q' \leftarrow Q' \cup T$
    \ENDIF
    \STATE $i \leftarrow i + 1$
\ENDWHILE
\RETURN $(Q', S)$
\end{algorithmic}
\label{alg}
\end{algorithm}

\subsection{Adversarial enhancement for mitigating transient blindness}
\label{smend}
Our analysis of tracing results reveals a post-edited model that exhibits excessive reliance on textual inputs. 
To address this modality imbalance, we leverage recent advancements in multimodal equilibrium techniques \citep{pi2025strengthening,chen2024we,wu2024noiseboost} by introducing an adversarial sample augmentation approach with regularization constraints. 
This method effectively rebalances image-text information contributions during model editing.

We adopt MEND \citep{mend}, a hypernetwork-based model editing method, as our baseline. As illustrated in Figure \ref{baseline}, MEND computes parameter updates by decomposing gradients of the edit loss $\mathcal{L}_e=-logp_{\theta'}(y_e|x_e^i,x_e^t)$ into low-rank matrices via hyper-network transformations. 
This ensures the model outputs the corrected answer aa for the edited input pair $x_e^i,x_e^t)$. 
To preserve unrelated predictions, MEND employs a locality loss $\mathcal{L}_{loc}$, defined as:
$\mathcal{L}_{loc}=\textbf{KL}(p_{\theta}(\cdot\mid x_o^i,x_o^t)\mid\mid p_{\theta'}(\cdot\mid x_o^i,x_o^t)) +\textbf{KL}(p_{\theta}(\cdot\mid x_o^t)\mid\mid p_{\theta'}(\cdot\mid x_o^t))$,
which constrains outputs for unrelated inputs  $(x_o^i,x_o^t)$.

However, as we discuss above, MEND’s locality constraints fail to address cross-modal leakage, the inadvertent alteration of predictions for semantically related but distinct inputs. 
{To balance the contribution between image and text modalities in the locality constraint, we use a KL divergence loss to ensure that the model’s output distribution remains consistent before and after editing:
\begin{equation}
\mathcal{L}_{loc}^M = \textbf{KL}\left(p_{\theta}(\cdot \mid x) || p_{\theta'}(\cdot  \mid x)\right),
\end{equation}
where $\theta$ and $\theta'$ are the model parameters before and after editing, and $x=(x_*^t,x_*^i)$ denotes a multimodal input composed of the edited textual question $x_*^t$ and an unaltered image $x_*^i$. }

We employ this loss function with MEND to let the model learn during training how to use image features more effectively when textual information is unreliable or mismatched.
This regularization could strengthen the model's reliance on image tokens when textual inputs become unreliable and penalize prediction divergence through KL divergence, promoting effective visual feature utilization.
The final loss function is:
\begin{equation}
    \textrm{Loss}=\mathcal{L}_e+\mathcal{L}_{loc}+\mathcal{L}_{loc}^{M},
    \label{loss_fun}
\end{equation}

This approach enhances the model's ability to process identical images with varying textual inputs by encouraging it to rely more on image information. Consequently, it improves the model's sensitivity to image features and strengthens its perception of the image modality after editing.

\begin{figure*}
    \centering
    \includegraphics[width=1\linewidth]{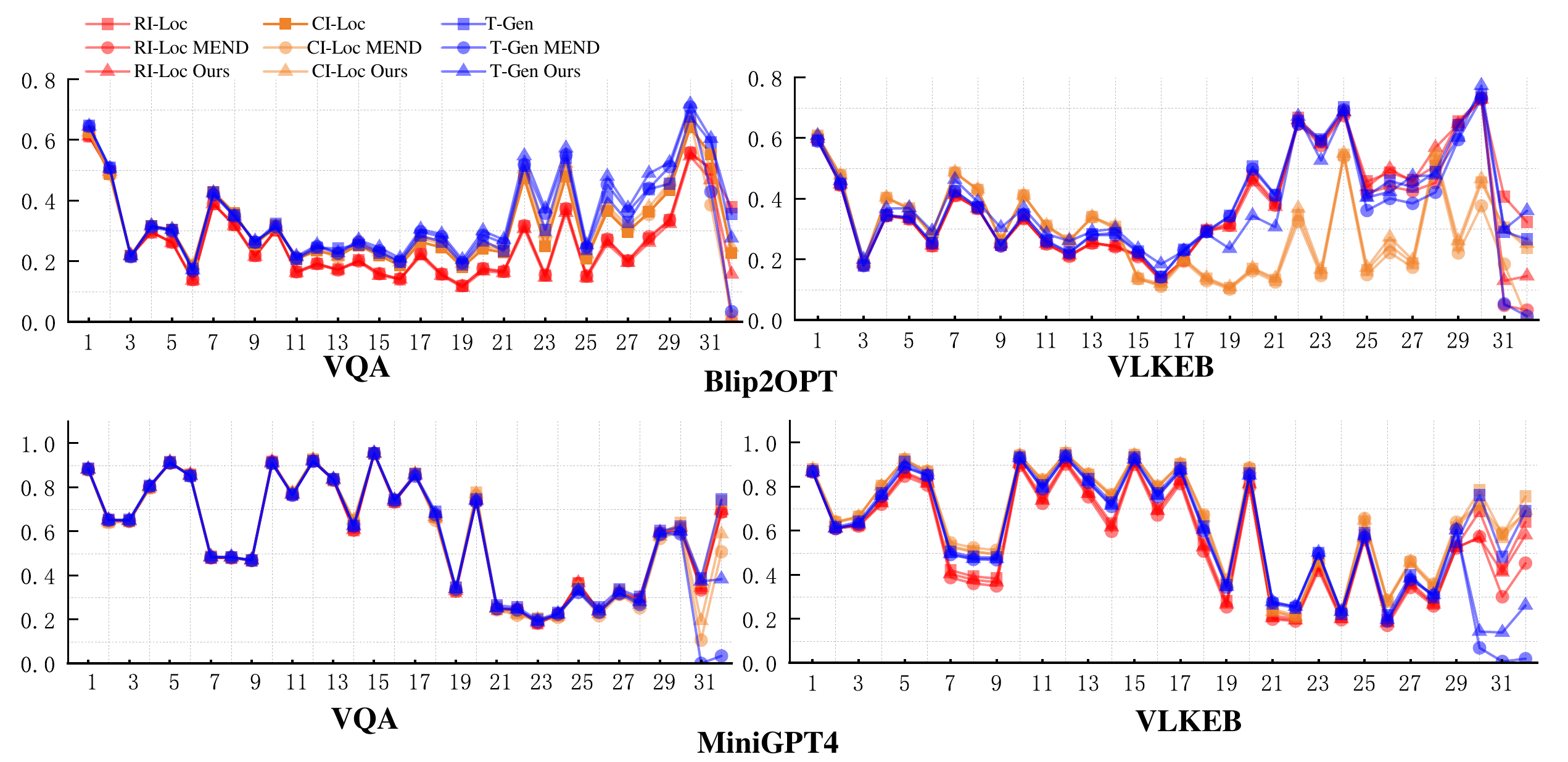}
    \caption{Image-text contribution to the model output.}
    \label{token_attributions_all}
\end{figure*}

\section{Results with De-VQA}
\label{typesres}
We provide detailed information on the different results here.
Table 3 presents a comparison of locality results across different types.
Table 4 shows the overall performance comparison of our method against others on DeVQA.
Tables 5 and 6 respectively report the ablation results for modules and loss functions.

In the results shown in Table 4, we observe that compared to the \{NI/RI/CI\}-Loc metric, the scores for the other three categories are significantly higher. This suggests that DeVQA can uncover limitations that are not captured by conventional evaluation approaches.
Our method achieves notable improvements on the NI, RI, and CI metrics while maintaining performance on the original evaluation criteria.
Combined with the results in Table 3, where our method achieves the best mean performance across different types of locality, these findings validate the effectiveness of our approach.

Table gives the results for Qwen-VL

\section{Editing Reduces Vision Token Contribution}
\label{tbs}
We calculated the ratio of contribution scores for image tokens versus text tokens at different layers when processing RI-Loc, CI-Loc, and T-Gen data:
\begin{equation}
Score^l=\frac{\sum_{i \in Q^l[image]}S^l[i]}{\sum_{i \in Q^l[text]}S^l[i]},
\label{scores}
\end{equation}
where $Q^l$ and $S^l$ are the key tokens and corresponding scores at layer $l$ in Algorithm \ref{alg}, respectively.
As shown in Figure \ref{token_attributions_all}, across the three different data types, we observed that in the edited layers (those beyond 29), both squares and triangles have higher values than circles, demonstrating that both our method and the pre-edited model can maintain the influence of image tokens during model prediction. 
In contrast, the MEND method significantly reduces the contribution of image tokens after editing, leading the model output to be predominantly driven by textual information.

Moreover, this phenomenon is more pronounced in higher layers. Below layer 20, the contributions of all token types are highly consistent, suggesting that the model primarily focuses on understanding the input during early stages. After layer 20, the model begins making decisions, allowing different types of tokens to play distinct roles. When editing is applied, it amplifies the influence of textual tokens while suppressing the contribution of visual tokens.
\begin{table}[h]
\centering
\begin{tabular}{ccccc}
\toprule
Model & Method & I-Loc    & T-Gen & CI-Loc   \\ \midrule 
\multirow{2}{*}{MiniGPT4} & MEND    & 0.6550  & 0.0372 & 0.0214 \\
 & Ours    & 0.6562   & 0.1956 & 0.2204 \\ \midrule
\multirow{2}{*}{Blip2OPT} &  MEND    & 0.4831  & 0.0788 & 0.0246  \\
 & Ours    & 0.5060  & 0.4699 & 0.3684 \\ \bottomrule
\end{tabular}
\caption{The ratio of the KL divergence (cf. Eq.(\ref{ration_eq})) between the image token representations and the text token representations before and after editing.}
\label{kls}
\end{table}

\textit{Existing methods that perform updates tend to primarily affect the representation of textual information, thereby diminishing the contribution of image information.}
To further explore why the model's reliance on image information is reduced, we computed the distribution of tokens in the model outputs (from the last layer of the LLM) before and after modifications and employed the Kullback-Leibler (KL) divergence to measure the disparity:
\begin{equation}
    Ratio=\frac{KL(P_{image}^{before}||P_{image}^{after})}{KL(P_{text}^{before}||P_{text}^{after})}, 
    \label{ration_eq}
\end{equation}
where $P_{image}$ and $P_{text}$ represent the image and text probability distributions in the model's output, respectively.

The results are shown in Table \ref{kls}. 
For I-Loc data, both MEND and our method affect the representations of image and text modalities after editing. 
However, for CI-Loc and T-Gen, the ratio of the KL divergence between the image token representations and the text token representations in MEND is significantly lower, indicating that the image token representations change much less compared to text token representations before and after editing.
In contrast, our method achieves a balanced update of both image and text tokens, thereby mitigating transient blindness.

\begin{table}[t]

\centering

\resizebox{.99\columnwidth}{!}{
\begin{tabular}{ccccc}
\toprule
Model                     & Metrics  & w/o    & w/     &  +Acc(\%) \\
\midrule
\multirow{5}{*}{QwneVL}   & In-Scope & 0.9474 & 0.9078 & -3.96\%                  \\
                          & T-loc    & 0.9757 & 0.9794 & 0.36\%                   \\
                          & I-loc    & 0.7599 & 0.7842 & 2.43\%                   \\
                          & RI-loc   & 0.195 & 0.5046 & 15.37\%                  \\
                          & CI-Loc   & 0.099 & 0.5151 & 15.35\%       \\ \bottomrule             
\end{tabular}}
\caption{Each metric represents the average performance after modifications to different modules.
 In-Scope is the average performance on Rel, T-Gen and I-Gen.
`w/' indicates with $L_{loc}^U$, `w/o' indicates without loss. }
\label{meanres}
\end{table}

\begin{table*}[]
\centering
\renewcommand\arraystretch{0.98}{
\begin{tabular}{cc|ccc|cc|cc|c|c|c}
\toprule
                        &          & \multicolumn{3}{c|}{CI-Loc} & \multicolumn{2}{c|}{NI-Loc} & \multicolumn{2}{c|}{RI-Loc} & T-Loc  & I-Loc  &        \\ \midrule
                       & Setting  & $T_1I_2$   & $T_2I_1$   & $T_2I_2$   & $T_1I_4$   & $T_2I_4$   & $T_1I_3$   & $T_3I_1$   & $T_3I_3$   & $T_4I_4$   & Mean   \\ \midrule
\multicolumn{12}{c}{Blip2OPT}                                                                                               \\ \midrule
\multirow{6}{*}{VLKEB} & MEND     & 0.4208 & 0.1809 & 0.4306 & 0.9950 & 0.4091 & 0.1288 & 0.6609 & 0.7516 & 0.9836 & 0.5513 \\ 
                       & LiveEdit & 0.0000 & 0.0000 & 0.0000 & 0.0000 & 0.0000 & 0.0000 & 0.6051 & 0.5861 & 1.0000 & 0.2435 \\
                       & RECIPE   & 0.0243 & 0.0224 & 0.0243 & 0.0243 & 0.0224 & 0.0243 & 0.6512 & 0.6985 & 1.0000 & 0.2769 \\
                       & LTE      & 0.2671 & 0.2761 & 0.2671 & 0.3256 & 0.3427 & 0.3053 & 0.6142 & 0.6845 & 0.9561 & 0.4487 \\
                       & TP       & 0.4811 & 0.4673 & 0.4811 & 0.4048 & 0.4030 & 0.4330 & 0.3189 & 0.4260 & 0.6794 & 0.4550 \\
        \rowcolor{gray!20}               & Ours     & 0.5155 & 0.4935 & 0.5186 & 1.0000 & 0.5085 & 0.5022 & 0.6698 & 0.7220 & 0.9788 & 0.6565 \\ \midrule
\multirow{7}{*}{VQA}   & MEND     & 0.0600 & 0.0150 & 0.5865 & 0.0400 & 0.1850 & 0.0700 & 0.6874 & 0.7557 & 0.9918 & 0.3768 \\
                       & LiveEdit & 0.1141 & 0.2630 & 0.0864 & 0.0000 & 0.0000 & 0.0000 & 0.6848 & 1.0000 & 1.0000 & 0.3498 \\
                       & RECIPE   & 0.2595 & 0.2352 & 0.2352 & 0.2595 & 0.2352 & 0.2595 & 0.6123 & 0.7973 & 1.0000 & 0.4326 \\
                       & LTE      & 0.2134 & 0.2937 & 0.6134 & 0.2949 & 0.7692 & 0.3226 & 0.2890 & 0.7848 & 0.9507 & 0.5035 \\
                       & TP       & 0.1822 & 0.2851 & 0.5822 & 0.2226 & 0.6302 & 0.3845 & 0.1053 & 0.4414 & 0.6663 & 0.3889 \\
                       & HICE     & 0.4940 & 0.4940 & 0.4940 & 0.4940 & 0.4940 & 0.4940 & 0.4940 & 0.5213 & 0.7455 & 0.5250 \\
        \rowcolor{gray!20}               & Ours     & 0.2500 & 0.3650 & 0.7643 & 0.3750 & 0.6000 & 0.4850 & 0.6948 & 0.8049 & 0.9686 & 0.5897 \\ \midrule
\multicolumn{12}{c}{MiniGPT4}                                                                                               \\ \midrule
\multirow{5}{*}{VLKEB} & MEND     & 0.2710 & 0.1843 & 0.2754 & 1.0000 & 0.2767 & 0.0541 & 0.7714 & 0.8169 & 0.9892 & 0.5154 \\
                       & LiveEdit & 0.0010 & 0.0054 & 0.0010 & 0.0000 & 0.0000 & 0.0000 & 0.6854 & 0.7512 & 1.0000 & 0.2716 \\
                       & RECIPE   & 0.0094 & 0.0041 & 0.0094 & 0.0094 & 0.0041 & 0.0094 & 0.7034 & 0.7931 & 1.0000 & 0.2825 \\
                       & TP       & 0.2841 & 0.3488 & 0.3841 & 0.3452 & 0.3497 & 0.3810 & 0.3981 & 0.5108 & 0.7385 & 0.4156 \\
        \rowcolor{gray!20}            & Ours     & 0.3575 & 0.3321 & 0.4767 & 1.0000 & 0.3900 & 0.4203 & 0.7591 & 0.7681 & 0.9906 & 0.6105 \\ \midrule
\multirow{6}{*}{VQA}   & MEND     & 0.0100 & 0.0250 & 0.8282 & 0.0850 & 0.2300 & 0.0300 & 0.7401 & 0.8873 & 0.9982 & 0.4260 \\
                       & LiveEdit & 0.1161 & 0.6178 & 0.1161 & 0.0000 & 0.0000 & 0.0050 & 0.6570 & 0.7451 & 1.0000 & 0.3619 \\
                       & RECIPE   & 0.4483 & 0.2901 & 0.4483 & 0.4483 & 0.2901 & 0.4483 & 0.6995 & 0.7891 & 1.0000 & 0.5402 \\
                       & TP       & 0.4299 & 0.5977 & 0.6299 & 0.4309 & 0.3824 & 0.5408 & 0.5928 & 0.7567 & 0.8661 & 0.5808 \\
                       & HICE     & 0.4940 & 0.4940 & 0.4940 & 0.4940 & 0.4940 & 0.4940 & 0.4940 & 0.5213 & 0.7773 & 0.5285 \\
   \rowcolor{gray!20}                    & Ours     & 0.5200 & 0.6450 & 0.8135 & 0.7600 & 0.8800 & 0.6200 & 0.6838 & 0.6903 & 0.9936 & 0.7340\\ \bottomrule
\end{tabular}}
\caption{Performance comparison of different Locality metrics.}
\label{TIs}
\end{table*}

\begin{table*}[]
\centering
\begin{tabular}{cccccccccccc}
\toprule
Method & FT     & FT-V   & SERAC  & MEND   & liveedit & recipe & lte & tp & lemoe & HICE    & Ours   \\ \midrule
\multicolumn{12}{c}{Blip2OPT   VQA}                                                                                  \\ \midrule
Rel    & 1      & 0.673  & 0.9964 & 0.999  & 0.9444   & 0.2453     & 1       & 0.8553 & 0.2353    & 0.964   & 0.9852 \\
T-Gen  & 0.9898 & 0.499  & 0.9978 & 0.9976 & 0.93     & 0.164      & 0.996   & 0.8464 & 0.1585    & 0.874   & 0.9718 \\
I-Gen  & 0.9898 & 0.499  & 0.9989 & 0.9976 & 0.8525   & 0.2453     & 1       & 0.8249 & 0.1961    & 0.974   & 0.9718 \\
T-Loc  & 0.7212 & 1      & 0.9967 & 0.9918 & 0.863    & 0.871      & 0.9507  & 0.6663 & 0.879     & 0.74547 & 0.9686 \\
I-Loc  & 0.1926 & 0.3387 & 0.0284 & 0.9557 & 0.734    & 0.8273     & 0.8848  & 0.4414 & 0.865     & 0.52125 & 0.8049 \\
NI-Loc & 0.2249 & 1      & 0.2785 & 0.04   & 0        & 0.2595     & 0.2949  & 0.2226 & 0         & 0.494   & 0.375  \\
RI-Loc & 0.0473 & 0.3326 & 0.0589 & 0.07   & 0        & 0.2595     & 0.3226  & 0.3845 & 0         & 0.494   & 0.585  \\
CI-Loc & 0.1393 & 0.7343 & 0.0036 & 0.5865 & 0.1141   & 0.2595     & 0.6134  & 0.5822 & 0         & 0.494   & 0.7643 \\ \midrule
\multicolumn{12}{c}{BLIP2OPT VLKEB}                                                                                  \\ \midrule
Rel    & 1      & 0.7022 & 0.9988 & 0.9881 & 0.9664   & 0.3732     & 1       & 0.5982 & 0.3821    &     -    & 0.9574 \\
T-Gen  & 0.9898 & 0.5878 & 0.9988 & 0.9821 & 0.9627   & 0.3873     & 0.9988  & 0.6082 & 0.3916    &     -    & 0.956  \\
I-Gen  & 0.9898 & 0.5878 & 0.9988 & 0.9821 & 0.8717   & 0.3732     & 1       & 0.5945 & 0.3926    &     -    & 0.956  \\
T-Loc  & 0.7212 & 1      & 1      & 0.9835 & 0.831    & 0.9        & 0.9561  & 0.6794 & 0.889     &     -    & 0.9788 \\
I-Loc  & 0.1926 & 0.3557 & 0.0241 & 0.7516 & 0.697    & 0.6985     & 0.7845  & 0.426  & 0.771     &     -    & 0.722  \\
NI-Loc & 0.2249 & 1      & 0.2973 & 0.995  & 0        & 0.0243     & 0.3256  & 0.4048 & 0         &     -    & 1      \\
RI-Loc & 0.0473 & 0.3549 & 0.0273 & 0.1288 & 0        & 0.0243     & 0.3053  & 0.433  & 0         &     -    & 0.5022 \\
CI-Loc & 0.1393 & 0.5176 & 0.3296 & 0.4306 & 0        & 0.0243     & 0.2671  & 0.4811 & 0         &     -    & 0.6186 \\ \midrule
\multicolumn{12}{c}{MiniGPT4 VQA}                                                                                    \\ \midrule
Rel    & 0.9689 & 0.3891 &    -    & 0.9971 & 0.898    & 0.4673     &     -    & 0.1979 & 0.2502    & 0.964   & 0.9943 \\
T-Gen  & 0.9819 & 0.2782 &    -    & 0.9952 & 0.886    & 0.4223     &     -    & 0.1948 & 0.2496    & 0.938   & 0.9871 \\
I-Gen  & 0.9521 & 0.2782 &    -    & 0.9952 & 0.7725   & 0.4673     &     -    & 0.2008 & 0.2973    & 0.992   & 0.9871 \\
T-Loc  & 0.8615 & 1      &    -    & 0.9982 & 0.951    & 0.912      &     -    & 0.8661 & 0.871     & 0.77733 & 0.9936 \\
I-Loc  & 0.2791 & 0.3432 &    -    & 0.8873 & 0.861    & 0.799      &     -    & 0.7567 & 0.761     & 0.52127 & 0.8903 \\
NI-Loc & 0.5392 & 1      &    -    & 0.085  & 0        & 0.4483     &     -    & 0.4309 & 0         & 0.494   & 0.76   \\
RI-Loc & 0.27   & 0.3398 &    -    & 0.03   & 0.005    & 0.4483     &     -    & 0.5408 & 0         & 0.494   & 0.62   \\
CI-Loc & 0.7915 & 0.7066 &    -    & 0.8282 & 0.1161   & 0.4483     &     -    & 0.6299 & 0         & 0.494   & 0.8135 \\ \midrule
\multicolumn{12}{c}{MiniGPT4 VLKEB}                                                                                  \\ \midrule
Rel    & 1      & 0.8098 &    -    & 0.9889 & 0.8512   & 0.4817     &     -    & 0.3077 & 0.4261    &     -    & 0.954  \\
T-Gen  & 0.9982 & 0.7628 &    -    & 0.9899 & 0.8607   & 0.4921     &     -    & 0.3141 & 0.4443    &     -    & 0.9711 \\
I-Gen  & 0.9982 & 0.7628 &    -    & 0.9899 & 0.7956   & 0.4817     &     -    & 0.3097 & 0.4278    &     -    & 0.9711 \\
T-Loc  & 0.8819 & 1      &    -    & 0.9892 & 0.931    & 0.941      &     -    & 0.7385 & 0.913     &     -    & 0.9906 \\
I-Loc  & 0.3454 & 0.4097 &    -    & 0.8169 & 0.716    & 0.775      &     -    & 0.5108 & 0.812     &     -    & 0.8681 \\
NI-Loc & 0.0044 & 1      &    -    & 1      & 0        & 0.0094     &     -    & 0.3452 & 0         &     -    & 1      \\
RI-Loc & 0.0627 & 0.4062 &    -    & 0.0541 & 0        & 0.0094     &     -    & 0.381  & 0         &     -    & 0.6203 \\
CI-Loc & 0.0139 & 0.4694 &    -    & 0.2754 & 0.001    & 0.0094     &     -    & 0.3841 & 0         &     -    & 0.5767 \\ \bottomrule
\end{tabular}
\caption{Results across all metrics. `-' means we could get the results do to the GPU memory.}
\end{table*}

\begin{table*}[]
\centering
\begin{tabular}{ccccc|cccc}
\toprule
     & MEND   & Ours   & MEND-DV & MEND-V & MEND   & Ours   & MEND-DV & MEND-V \\ \midrule
     & \multicolumn{4}{c|}{VLKEB}          & \multicolumn{4}{c}{VQA}            \\ \midrule 
     & \multicolumn{8}{c}{Blip2OPT}                                            \\ \midrule
$T_4I_4$ & 0.9836 & 0.9788 & 0.9854  & 1      & 0.9918 & 0.9686 & 0.9917  & 1      \\
$T_3I_3$ & 0.7516 & 0.722  & 0.7861  & 0.4056 & 0.9557 & 0.8049 & 0.9407  & 0.5342 \\
$T_3I_1$ & 0.6609 & 0.6698 & 0.6719  & 0.233  & 0.6847 & 0.6948 & 0.7642  & 0.1218 \\
$T_1I_2$ & 0.4208 & 0.5155 & 0.4303  & 0.6174 & 0.06   & 0.25   & 0.045   & 0.54   \\
$T_1I_3$ & 0.1288 & 0.5022 & 0.1339  & 0.4824 & 0.07   & 0.485  & 0.04    & 0.685  \\
$T_1I_4$ & 0.995  & 1      & 1       & 1      & 0.04   & 0.375  & 0.02    & 1      \\
$T_2I_1$ & 0.1809 & 0.4935 & 0.2042  & 0.4633 & 0.015  & 0.365  & 0.01    & 0.01   \\
$T_2I_2$ & 0.4306 & 0.5186 & 0.4398  & 0.6253 & 0.5865 & 0.7643 & 0.559   & 0.6025 \\
$T_2I_4$ & 0.4091 & 0.5085 & 0.4304  & 0.6476 & 0.185  & 0.6    & 0.115   & 1      \\
MEAN & 0.5513 & 0.6566 & 0.5647  & 0.6083 & 0.3987 & 0.5897 & 0.3873  & 0.6103 \\ \midrule
     & \multicolumn{8}{c}{MiniGPT4}                                            \\ \midrule
$T_4I_4$ & 0.9892 & 0.9906 & 0.9904  & 1      & 0.9982 & 0.9936 & 0.9923  & 1      \\
$T_3I_3$ & 0.8169 & 0.7681 & 0.8077  & 0.5132 & 0.8873 & 0.6903 & 0.8964  & 0.66   \\
$T_3I_1$ & 0.7714 & 0.7591 & 0.7059  & 0.3789 & 0.8401 & 0.6838 & 0.8445  & 0.2013 \\
$T_1I_2$ & 0.271  & 0.3575 & 0.2485  & 0.4228 & 0.01   & 0.52   & 0.005   & 0.565  \\
$T_1I_3$ & 0.0541 & 0.3203 & 0.0572  & 0.2901 & 0.03   & 0.62   & 0.02    & 0.945  \\
$T_1I_4$ & 1      & 1      & 1       & 1      & 0.085  & 0.76   & 0.06    & 0.98   \\
$T_2I_1$ & 0.1843 & 0.3321 & 0.0963  & 0.1214 & 0.025  & 0.645  & 0.035   & 0.015  \\
$T_2I_2$ & 0.2754 & 0.3767 & 0.2463  & 0.4546 & 0.8282 & 0.8135 & 0.8363  & 0.855  \\
$T_2I_4$ & 0.2767 & 0.39   & 0.2578  & 0.4842 & 0.23   & 0.88   & 0.205   & 0.975  \\
MEAN & 0.5154 & 0.5883 & 0.49    & 0.5184 & 0.4371 & 0.734  & 0.4327  & 0.6885 \\ \bottomrule
\end{tabular}
\caption{Results for editing different Modules.}
\end{table*}

\begin{table*}[]
\centering
\begin{tabular}{ccccccccc}
\toprule
Setting & MEND     & NI       & RI       & CI       & CI-NI    & CI-RI    & RI-NI    & CI-RI-NI \\ \midrule
        & \multicolumn{8}{c}{Blip2OPT VLKEB}                                                    \\ \midrule
$T_4I_4$    & 0.9836   & 0.9611   & 0.9611   & 0.9604   & 0.9722   & 0.9705   & 0.9801   & 0.9788   \\
$T_3I_3$    & 0.7516   & 0.5003   & 0.5202   & 0.501    & 0.581    & 0.569    & 0.7222   & 0.722    \\
$T_3I_1$    & 0.6609   & 0.4263   & 0.4457   & 0.4219   & 0.5079   & 0.4942   & 0.6475   & 0.6698   \\
$T_1I_2$    & 0.4208   & 0.5065   & 0.4996   & 0.4739   & 0.5213   & 0.4957   & 0.5172   & 0.5155   \\
$T_1I_3$    & 0.1288   & 0.4688   & 0.4748   & 0.4285   & 0.4927   & 0.4634   & 0.2151   & 0.5022   \\
$T_1I_4$    & 0.995    & 1        & 1        & 1        & 1        & 1        & 1        & 1        \\
$T_2I_1$    & 0.1809   & 0.4618   & 0.4701   & 0.4444   & 0.4885   & 0.4572   & 0.3437   & 0.4935   \\
$T_2I_2$    & 0.4306   & 0.5118   & 0.502    & 0.4745   & 0.5297   & 0.5021   & 0.5192   & 0.5186   \\
$T_2I_4$    & 0.4091   & 0.529    & 0.5172   & 0.4765   & 0.5193   & 0.5046   & 0.5748   & 0.5085   \\
MEAN    & 0.5513   & 0.5962   & 0.599    & 0.5757   & 0.6236   & 0.6063   & 0.6133   & 0.6566   \\ \midrule
        & \multicolumn{8}{c}{Blip2OPT VQA}                                                      \\ \midrule
$T_4I_4$    & 0.9918   & 0.935    & 0.9557   & 0.9617   & 0.9632   & 0.9491   & 0.9973   & 0.9686   \\
$T_3I_3$    & 0.9557   & 0.7591   & 0.7511   & 0.7668   & 0.7758   & 0.7537   & 0.9277   & 0.8049   \\
$T_3I_1$    & 0.6847   & 0.6624   & 0.6082   & 0.6175   & 0.6875   & 0.6298   & 0.6052   & 0.6948   \\
$T_1I_2$    & 0.06     & 0.155    & 0.075    & 0.09     & 0.305    & 0.085    & 0.455    & 0.25     \\
$T_1I_3$    & 0.07     & 0.36     & 0.2      & 0.23     & 0.525    & 0.22     & 0.98     & 0.485    \\
$T_1I_4$    & 0.04     & 0.265    & 0.15     & 0.19     & 0.46     & 0.175    & 1        & 0.375    \\
$T_2I_1$    & 0.015    & 0.265    & 0.205    & 0.22     & 0.405    & 0.215    & 0.03     & 0.365    \\
$T_2I_2$    & 0.5865   & 0.7299   & 0.6349   & 0.6448   & 0.7881   & 0.6549   & 0.6959   & 0.7643   \\
$T_2I_4$    & 0.185    & 0.54     & 0.405    & 0.425    & 0.66     & 0.43     & 1        & 0.6      \\
MEAN    & 0.3987   & 0.5191   & 0.4428   & 0.4606   & 0.6188   & 0.4569   & 0.7435   & 0.5897   \\ \midrule
        & \multicolumn{8}{c}{MiniGPT4 VLKEB}                                                    \\ \midrule
$T_4I_4$    & 0.9892   & 0.9878   & 0.9899   & 0.9898   & 0.9875   & 0.9901   & 0.9852   & 0.9906   \\
$T_3I_3$    & 0.8169   & 0.7201   & 0.7394   & 0.7391   & 0.7311   & 0.755    & 0.7808   & 0.7681   \\
$T_3I_1$    & 0.7714   & 0.7124   & 0.7249   & 0.7247   & 0.7226   & 0.7417   & 0.7297   & 0.7591   \\
$T_1I_2$    & 0.271    & 0.2757   & 0.2125   & 0.2349   & 0.2777   & 0.2461   & 0.3781   & 0.3575   \\
$T_1I_3$    & 0.0541   & 0.2332   & 0.1396   & 0.1653   & 0.2191   & 0.1839   & 0.0541   & 0.3203   \\
$T_1I_4$    & 1        & 1        & 1        & 1        & 1        & 1        & 1        & 1        \\
$T_2I_1$    & 0.1843   & 0.2534   & 0.1766   & 0.1997   & 0.2416   & 0.2047   & 0.208    & 0.3321   \\
$T_2I_2$    & 0.2754   & 0.2983   & 0.2208   & 0.2471   & 0.2886   & 0.2596   & 0.3708   & 0.3767   \\
$T_2I_4$    & 0.2767   & 0.3503   & 0.2632   & 0.2933   & 0.3431   & 0.2936   & 0.4778   & 0.39     \\
MEAN    & 0.5154   & 0.5368   & 0.4963   & 0.5104   & 0.5346   & 0.5194   & 0.5538   & 0.5883   \\ \midrule
        & \multicolumn{8}{c}{MiniGPT4 VQA}                                                      \\ \midrule
$T_4I_4$    & 0.9982   & 0.9964   & 0.9705   & 0.9711   & 0.9945   & 0.9952   & 0.9713   & 0.9936   \\
$T_3I_3$    & 0.8873   & 0.6484   & 0.6253   & 0.6441   & 0.6794   & 0.6587   & 0.8315   & 0.6903   \\
$T_3I_1$    & 0.8401   & 0.6436   & 0.6123   & 0.6127   & 0.6685   & 0.6447   & 0.6421   & 0.6838   \\
$T_1I_2$    & 0.01     & 0.52     & 0.495    & 0.485    & 0.505    & 0.485    & 0.385    & 0.52     \\
$T_1I_3$    & 0.03     & 0.605    & 0.55     & 0.505    & 0.585    & 0.535    & 0.9      & 0.62     \\
$T_1I_4$    & 0.085    & 0.67     & 0.55     & 0.51     & 0.675    & 0.55     & 0.96     & 0.76     \\
$T_2I_1$    & 0.025    & 0.62     & 0.6      & 0.545    & 0.62     & 0.6      & 0.02     & 0.645    \\
$T_2I_2$    & 0.8282   & 0.8231   & 0.8153   & 0.8187   & 0.8185   & 0.8193   & 0.8243   & 0.8135   \\
$T_2I_4$    & 0.23     & 0.82     & 0.71     & 0.625    & 0.855    & 0.705    & 0.97     & 0.88     \\
MEAN    & 0.4371   & 0.7052   & 0.6587   & 0.6352   & 0.7112   & 0.6659   & 0.7227   & 0.734    \\ \bottomrule
\end{tabular}
\caption{Results for editing with different loss combination.}
\end{table*}

\end{document}